\documentclass{article}

\usepackage[preprint]{neurips_2024}

\usepackage[utf8]{inputenc} 
\usepackage[T1]{fontenc}    
\usepackage{hyperref}       
\usepackage{url}            
\usepackage{booktabs}       
\usepackage{amsfonts}       
\usepackage{nicefrac}       
\usepackage{microtype}      
\usepackage{xcolor}         
\usepackage{graphicx}       

\graphicspath{ {./images/} }

\title{Drama Engine: A Framework for Narrative Agents}

\author{%
  Martin Pichlmair\\
  Write with LAIKA\\
  Copenhagen, Denmark\\
  \texttt{martin@writewithlaika.com} \\
  \And
  Riddhi Raj\\
  Write with LAIKA\\
  Copenhagen, Denmark\\
  \texttt{riddhi@writewithlaika.com} \\
  \AND
  Charlene Putney\\
  Write with LAIKA\\
  Copenhagen, Denmark\\
  \texttt{char@writewithlaika.com} \\
}

\begin{document}

\maketitle

\begin{abstract}
This technical report presents the Drama Engine\footnote{Code: \url{https://github.com/Write-with-LAIKA/drama-engine}}, a novel framework for agentic interaction with large language models designed for narrative purposes. The framework adapts multi-agent system principles to create dynamic, context-aware companions that can develop over time and interact with users and each other. Key features include multi-agent workflows with delegation, dynamic prompt assembly, and model-agnostic design. The Drama Engine introduces unique elements such as companion development, mood systems, and automatic context summarising. It is implemented in TypeScript. The framework's applications include multi-agent chats and virtual co-workers for creative writing. The paper discusses the system's architecture, prompt assembly process, delegation mechanisms, and moderation techniques, as well as potential ethical considerations and future extensions.
\end{abstract}

\section{Introduction}

Autonomous agents are currently one of the most dynamic areas of applied machine learning. Several frameworks for running agents have been published in recent years, e.g. AutoGen \citep{wuAutoGenEnablingNextGen2023}, CrewAI\footnote{\url{https://www.crewai.com}}, and AgentGPT\footnote{\url{https://agentgpt.reworkd.ai}}. \citep{wangSurveyLargeLanguage2024} and \citep{xiRisePotentialLarge2023} provide very complete overviews of the state of research. Most frameworks focus on single agents that autonomously solve problems in multi-step processes using a large language model as the back-end. They are implemented as Python libraries that run on the command line or servers. Some feature web-based front-ends or configuration interfaces. Some use multi-agent conversations to solve a problem. In the simplest case, one agent has the ability to write code, another agent has the ability to run code and a third agent has the ability to provide feedback to the first agent based on the results of running the code.

While these multi-agent systems internally run a conversation with several participants, their intended use is to come up with a solution to a problem specified by the user. The user’s next interaction would be critique of said solution and so forth. Apart from the orchestration of the interaction of agents with each other and with the user, the main innovation in this field is dynamic prompt assembly, e.g. by adding memory to the agents \citep{packerMemGPTLLMsOperating2023, zhangSurveyMemoryMechanism2024, zhengMemoryRepositoryAINPC2024} or by incorporating the feedback from another agent when rewriting a prompt \citep{wuAutoGenEnablingNextGen2023}.

With the Drama Engine, we have adapted the basic principles of multi-agent systems for narrative purposes, see also \cite{huSurveyLargeLanguage2024a}. We have created a system that takes established methods of how companions are realised in video games and uses those for multi-companion chats as well as single-companion chats. Unlike \citep{parkGenerativeAgentsInteractive2023a} and \citep{yaoKeepCALMExplore2020} we focus on dialog and role-play (see also \cite{nananukulWhatIfRed2024, urbanekLearningSpeakAct2019a, wangRoleLLMBenchmarkingEliciting2023, mehtaExploringViabilityConversational2022}). 

This document serves two purposes: to explain the reasoning behind design decisions and to serve as a manual for the Drama Engine.

\section{The Drama Engine Framework}

The Drama Engine is a framework for agentic interaction with language models. It is written in TypeScript. It runs in any modern JavaScript engine including browsers and Node.

At the heart of the drama engine are companions and their orchestration. Some companions - those with which the user interacts directly - are agents that simulate a personality. They can change over time, have moods, and interact with each other. These companions use deputies to run ad-hoc chains of prompts that allow for a mix of different prompting techniques. A deputy might use a few-shot prompt to an instruction-tuned model while its host companion talks to the user by calling a chat-tuned model. This way, dynamic sequences of prompting (e.g. text summary, but only if the text is too long $\rightarrow$ text analysis $\rightarrow$ discussion about the analysis) can be configured in a modular way. The resulting system is far more flexible than plain prompt chains and more controllable than pure agentic solutions.

The Drama Engine has the following core features:
\begin{itemize}
  \item \textbf{Multi-agent workflows with delegation:} The conversation between several agents is orchestrated via a moderator. Agents can delegate more complex tasks to chains of deputies.
  \item \textbf{Dynamic prompt assembly:} The prompt sent to the back-end is assembled based on context. This context includes data about other chat participants, the task, the state of the agent, the materials necessary to complete the task, and so on. Details below.
  \item \textbf{Model- and vendor-agnostic:} When run locally, the Drama Engine can use any back-end that supports OpenAI’s API standard. We have tested it with Together AI's services The framework works with any language model. It supports ChatML and Mistral prompt formats. We are using the framework with teknium’s \texttt{OpenHermes-2p5-Mistral-7B}\footnote{\url{https://huggingface.co/teknium/OpenHermes-2.5-Mistral-7B}} and Mistral’s \texttt{Mixtral-8x7B-Instruct-v0.1}\footnote{\url{https://huggingface.co/mistralai/Mixtral-8x7B-Instruct-v0.1}} in production.
\end{itemize}

There are some unique features to the Drama Engine that no other system supports in this way. 
\begin{itemize}
  \item \textbf{Companions develop over time:} In video games it is normal that new conversations with your party members get unlocked as the adventure progresses. We’ve added similar functionality to the companion configuration so you can set conditions that have to be met. Those can be based on stats tracked inside the Drama Engine, e.g. number of interactions with the companion, or based on external stats that come from the game world.
  \item \textbf{Companions have moods:} At configuration time, moods and their effects on the prompt as well as probabilities of occurrence can be specified.
  \item \textbf{Automatic summaries of context:} The system is meant for having the companions work with external context provided by whatever the Drama Engine is connected to. In the case of our own Writers Room, that is the text the user is writing. If the amount of supplied data exceeds the context size of the model the system can automatically summarise the data using a number of methods.
  \item \textbf{Environment reactions:} The framework is intended to be used in an environment that supplies additional stats that the companions can react to. This is done via a collection of key-value pairs that are exposed to the prompt assembler.
\end{itemize}

The unique features of the Drama Engine are mostly translations of existing concepts of how video game characters are developed to an agentic flow. We aim for a balance between the predictability of the companions and the freedom LLMs offer for dialogues.

\subsection{Companions}

\begin{table}
  \caption{Companion Configuration}
  \label{companionconfig-table}
  \centering
  \begin{tabular}{lll}
    \toprule
    Field name          & Type                      & Description \\
    \midrule
    \multicolumn{3}{l}{\textit{Data fields}}\\
    \midrule
    \texttt{name}       & \texttt{string}           & The name of the companion\\
    \texttt{class}      & \texttt{AutoCompanion}    & \texttt{AutoCompanion} sub-class to be instantiated\\
    \texttt{description}   & \texttt{string}         & Machine-readable description\\
    \texttt{base\_prompt}   & \texttt{string}         & The base prompt of the companion\\
    \texttt{kind}       & \texttt{CompanionKind}     & \texttt{"user" | "npc" | "shell"}\\
    \midrule
    \multicolumn{3}{l}{\textit{Data fields (optional)}}\\
    \midrule
    \texttt{bio}       & \texttt{string}           & Human-readable bio\\
    \texttt{avatar}       & \texttt{string}       & Link to an avatar image\\
    \texttt{job}       & \texttt{string}       & Base job\\
    \texttt{situations}       & \texttt{LIST}       & A list of prompt pieces for each situation\\
    \texttt{knowledge}       & \texttt{LIST}      & A list of conditional lines of knowledge\\
    \texttt{mottos}       & \texttt{LIST}      & A list of conditional lines of mottoes\\
    \texttt{moods}       & \texttt{LIST}      & A list of moods and their probabilities and prompt pieces\\
    \midrule
    \multicolumn{3}{l}{\textit{Actions and triggers (optional)}}\\
    \midrule
    \texttt{actions}       & \texttt{LIST}      & A list of action descriptions\\
    \texttt{triggers}       & \texttt{LIST}      & A list of trigger descriptions (WIP)\\
    \texttt{scope}       & \texttt{LIST}      & The application-related scope a deputy works in\\
    \midrule
    \multicolumn{3}{l}{\textit{Model configuration (optional)}}\\
    \midrule
    \texttt{modelConfig}       & \texttt{ModelConfig}      & Override all default model parameters\\
    \texttt{temperature}       & \texttt{number}      & Override the inference temperature\\
    \bottomrule
  \end{tabular}
\end{table}

There are three kinds of companions in the Drama Engine: User, NPC and shell. They are all sub-classes of \texttt{AutoCompanion} which is the only subclass of the abstract class Companion.

There is always only one user. NPCs (non-playing characters) are companions the user talks to directly. Shell companions are mostly deputies to the NPCs that offer specialised functionalities. The only other shell companion is the moderator of the chat.
The configuration of companions is a JSON structure and provided to the Drama Engine at startup. The structure also defines which class is used to instantiate the companion. NPCs should have the class \texttt{ChatCompanion}. Most simple shells will be \texttt{InstructionDeputy} instances. Specialised deputies can be written as new sub-classes of \texttt{Deputy}.

Agents are configured in an instance of \texttt{CompanionConfig} that defines their identities. It covers their names, backstories, and how they develop over time – everything that defines their behaviour. Table \ref{companionconfig-table} gives a full overview of the configurable parameters. Different types of companions use different fields. The fields have different data types, most of them either being strings or lists of structures.

\subsubsection{NPCs}

NPCs are the characters the user chats with. Additionally to the mandatory data fields they access moods, the bio, and many other fields. Here is an overview.

The bio is written for the user. It is inspired by social media bios. The description is used by the moderator and other companions to know who is who. The base prompt is the first part of every prompt the companions sends to the model so it should start with their name and then have a few lines of definitions of how the character is supposed to talk and who they are. Situations are defined per chat and the prompt in them is added to every prompt sent to the model that happens in a chat room that plays in that situation.

Moods are probabilistic; you define a set of moods and how probable they are. If the sum of all probabilities adds up to less than 1, the character is in a “neutral” mood if the dice rolls a value in that range. Moods are currently initialised on load – which means they change on reload.

Knowledge unlocks as the user interacts with the companions or other tracked values of the system. This mechanism establishes a feeling of growing familiarity as the companion opens up more and more to the user.

\subsubsection{Shells}

This type of companion never talks directly to the user. They are employed by NPCs to execute various actions or used to orchestrate the chat. The chat moderator is preconfigured. The simplest deputy is the InstructionDeputy. This deputy defines a single \texttt{replyFunction}. There are more reply functions hooked up in Deputy, the super-class of \texttt{InstructionDeputy} but we’ll get to that later. A reply function triggers when the deputy gets activated and certain conditions are met (see below). The single reply function found in this file has a condition that is always met and adds the job defined in the companion’s configuration to the prompt.

The configuration of an \texttt{InstructionDeputy} should define at least their job (which gets added to the prompt for them to execute their abilities), a scope, and their name. The scope defines what part of the context the deputy acts on. Possible scopes are the last sentence, the last paragraph, a random paragraph, “some” (meaning “anything that’s not nothing"), or the full document. The name is only used to hook up actions. Deputies can set their temperature – most times a low value is best suited for these more structured inferences.

\subsection{Reply functions}

Reply functions are a way for a companion object to control the flow of execution. Every companion has a list of reply functions, and they get executed one by one until one of them returns true, meaning the companion is done with their job. They are currently used for three purposes. Firstly, all companions use the catch-all reply function to execute their base functionality. Secondly, if a provided text is too long for the context size of the model, all companions can use a set of summarising and scoping functions to boil down the context to its essence. Thirdly, actions that require input data will stop executing and tell the user to provide more data if there is too little.
The triggers for reply functions can take a lot of different shapes as evident in the \texttt{evaluateReplyTrigger} in the \texttt{AutoCompanion} class. Triggers will execute when a certain condition is met. Possible conditions that lead to execution of a reply function are that the specified value is the active action, the sender’s name, an instance of \texttt{AutoCompanion}, the result of a function call, or smaller than a random number. In the case of a function call that function receives the current context and the last speaker as its input and should return a boolean value. We use that to determine if a summary has to be made because the context contains all text and since the context is edited by all reply functions before being passed on, our summariser can do its whole job before the deputy or companion acts on the context.

\subsection{Context object and world state}

\begin{table}
  \caption{Context object fields}
  \label{context-table}
  \centering
  \begin{tabular}{ll}
    \toprule
    Field name     & Description \\
    \midrule
    \multicolumn{2}{l}{\textit{Downstream (user $\rightarrow$ companion $\rightarrow$ prompt)}}\\
    \midrule
    \texttt{chat} & The compressed chat history (only for the moderator \& care bot)\\
    \texttt{knowledge} & Unlocked knowledge\\
    \texttt{text} & A line of text provided by the client\\
    \texttt{paragraph} & A paragraph of text provided by the client\\
    \texttt{epilogue} & Text that should always go to the end of the system prompt\\
    \texttt{input} & An input provided by a user\\
    \texttt{action} & An active action\\
    \midrule
    \multicolumn{2}{l}{\textit{Midstream (companion $\rightarrow$ prompt)}}\\
    \midrule
    \texttt{persona} & The parts of the person that should be exhibited in this prompt\\
    \texttt{job} & The instruction to the LLM\\
    \texttt{mood} & Change the mood of the companion\\
    \midrule
    \multicolumn{2}{l}{\textit{Upstream (companion $\rightarrow$ user)}}\\
    \midrule
    \texttt{question} & A question by the deputy\\
    \texttt{answer} & An answer returned by a deputy\\
    \texttt{excerpt} & An item that should be added to the chat instead of the "message".\\
    \texttt{quote} & An item that should be added to the chat instead of a chat message\\
    \texttt{message} & The message to be added to the chat history. Also goes into the  prompt.\\
    \midrule
    \multicolumn{2}{l}{\textit{Internal}} \\
    \midrule
    \texttt{companionNames} & The other companions around, comma-separated\\
    \texttt{error} & An error that occurred\\
    \texttt{conversationID} & A conversation can have multiple turns. They all share an ID\\
    \midrule
    \multicolumn{2}{l}{\textit{Tool usage (not used yet)}}\\
    \midrule
    \texttt{tool} &  A tool to use\\
    \bottomrule
  \end{tabular}
\end{table}

The link between companions and the work of the user is established using two kinds of objects. The world state is a single global database of key-value pairs that can hold any kind of information of relevance to the system (e.g. USERNAME stores the user’s name in our own implementation so the companions can address them by their name). The context object gets constructed for each interaction of the user with the drama engine. It gets passed between companions and down to deputies. One context is valid for one exchange. Under some circumstances an exchange can have multiple turns of user interaction (i.e. when the companion asks a question and is waiting for an answer).

The \texttt{Context} class is well documented in the code. The actual act of prompting the language model packs the context and a model configuration (plus some administrative data) into a job that gets turned into a query for the model. That means all information needed to create the prompt is in the context. That information ranges from who is the speaker to what their job is at this moment, to all the data from the user, to who else is in the room, etc. Some information (e.g. the mood) is added by the companion before it passes the context on. If the companion delegates the action, it might also add information for the delegate. The delegate writes their reply into the context (potentially after an inference), and the companion reads it from there before acting on it.
Table \ref{context-table} shows an overview of all fields in the \texttt{Context} class.

\subsection{Prompt assembly}

All prompts are assembled on the fly depending on the context. This happens in the \texttt{assemblePrompt} function in the Prompter class. The prompter takes the Context object, the world state database and transforms them into a prompt using decorators and other mechanisms. Decorators are simple replacement-based templates that are used to tag specific pieces of information for referencing them in a job. E.g. we’re using the decorator \texttt{USER TEXT="\{\{DATA\}\}"} to label the user-provided textual data as \texttt{USER TEXT}, so a job can say “Summarise the USER TEXT” and the model will understand where to find the text in most cases. The prompter also adds some default information like the current data and time.

The final prompt format (e.g. ChatML) is applied in a second step where the model uses a template to convert the prompt data to the right format.

\subsection{Delegation and actions}

Our system mixes chained prompting with explicit so-called actions. An action is a concrete specific activity to be performed by the system. An example would be “Summarise my text”, which we offer in the Writers Room. In order to specify an action, there has to be a deputy that executes it. A deputy can be hooked up to any number of actions. Actions can have conditions, if they should unlock over time or only be available during specific circumstances. Actions are meant to be triggered explicitly.

If an action is triggered, the moderator automatically makes the deputy who is supposed to execute it the next speaker and the companion who hosts the action the one following up. The deputy either executes their own language model call or just defines the job of the companion and returns. By default the companion writes to the chat in place of the deputy. Depending on what field of the context the deputy writes their data to, the deputy’s output might additionally enter the chat. See \texttt{runChat} in the Drama object or Context for details.

New functionality is easiest to add to the Drama Engine by adding new deputies with specific abilities.

\subsection{Chats and moderation}

The Drama object orchestrates all chats. In our next refactoring we will transfer some of the functionality from that object to the Chat object but for now reading the Drama class should give a good overview of how chats work. The main entry point for a multi-round chat is the \texttt{runConversation} function. This function runs a conversation (via \texttt{runChat}) for up to a specified amount of rounds. Each round it calls the moderator to determine the next speaker. If the next speaker is the user, the conversation ends. Otherwise, it calls the \texttt{generateReply} function of one active speaker after the other passing the same context object along between them. If a reply should be appended to the chat messages, it does so. The \texttt{runChat} function also updates the state of all participants, counts the interactions, and keeps the chat database in sync.

The moderator selects the next speaker. The moderation function allows for a high level of customisation. The default speaker selection flow is described in \ref{multi-agent-chat} below. In addition, specific speakers can be excluded from the list of allowed speakers. Repeat replies from the same companion can also be disabled. Shells (deputies) are not permitted to speak except as part of an action.

\section{Applications of the Drama Engine}

The intended application of the Drama Engine is a case where a collection of companions has to use specific functionality to support the user with a task. Ideally, there is user-provided context in the form of text to work with. In the Writers Room we are using the Drama Engine for two tasks. First, we have a multi-agent chat where the conversation orchestration is using the moderation system of the Drama Engine. It’s also where unlockable knowledge of companions surfaces the most.

\subsection{A1: Multi-agent chat} \label{multi-agent-chat}

\begin{figure}
    \caption{The control flow of a multi-agent conversation orchestrated by the moderator}
    \centering
    \includegraphics[width=1.0\textwidth]{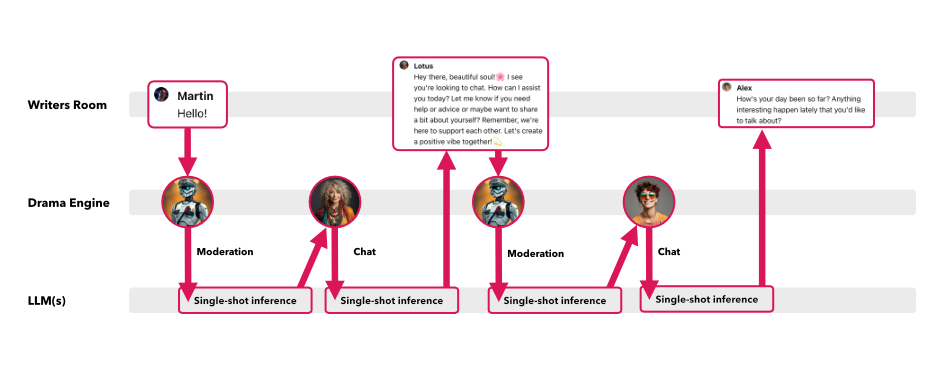}
    \label{fig:conversation}
\end{figure}

The multi-agent chat, shown in figure \ref{fig:conversation}, uses the multi-turn conversation interface of the Drama Engine. It runs a conversation for a random number of rounds larger than 3 and smaller the number of chat participants plus one. The conversation is always initiated by the user. The user can be returned as the next speaker if the moderator concludes that that is the right thing to do.

The next speakers are returned according to the following prioritisation:

\begin{enumerate}
\item A recipient or list of recipients in the current context (which usually means a deputy has to be activated)
\item The only other chat participant if it's a 1:1 chat
\item All chat participants mentioned in the last message in that order
\item If the speaker selection mode is round robin, the next speaker in the list
\item If the speaker selection is random, a random speaker
\item The recent chat history is collected and an LLM is prompted to pick the next logical speaker
\item As a fallback, a random speaker is returned
\end{enumerate}

After the next speakers are selected, the correct prompt is assembled based on the context. Especially the situation, the chat history, and the unlocked knowledge are relevant. All speakers in the list are prompted in sequence. Each reply is posted to the supplied callback before the next speaker is handled.

\subsection{A2: Virtual coworker for creative writing}

\begin{figure}
    \caption{The control flow of a deputised action}
    \centering
    \includegraphics[width=1.0\textwidth]{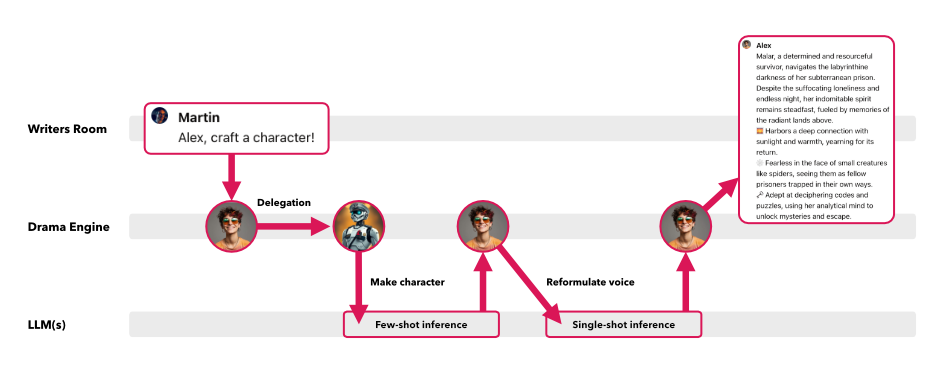}
    \label{fig:deputisation}
\end{figure}

The other application of the Drama Engine in the Writers Room is task-based interaction with companions. Figure \ref{fig:deputisation} shows the control flow of this operation. Here, two different ways of interacting with the companions are supported. In both cases, speaker selection uses the prioritisation listed above but always returns within the first two steps.

First, direct chat focused on a provided piece of data. In our case, the selected paragraph or the text visible on the screen are picked as data and supplied to the conversation via the Context object. This data gets automatically added to the prompt in a way that encourages the LLM to consider it in its reply. If the data exceeds the context window of the LLM, the text is automatically summarised. 

The second way of interacting with companions in the Writers Room is via so-called “actions”. These are deputy-based tasks that either use the generic Instruction Deputy or one of several custom deputies we created for the Writers Room. The latter are hosted in the Writers Room but sub-classes of the same Deputy class as the Instruction Deputy. An action is a simple mapping between a deputy and a companion that has a name and an ID so it can be represented in the interface. We render them as simple buttons that lead to a conversation with the companion. Once the button is pressed, a context is created that conforms to the expectations of the deputy. Here, some actions work with the full document, automatic summaries if the context window of the model is too small, or shorter pieces of data, depending on the purpose of the action. An action that finds the theme of the text in question has to work with the whole document, whereas an action for rewriting a line of dialog would only target the current sentence.

\section{Discussion}

The Drama Engine is a product and not a research project. It has been evaluated by being integrated into the Writers Room\footnote{\url{http://writers-room.com}}, our live product. We have conducted a series of user testing sessions to collect qualitative feedback about working with the Writers Room. Here are some observations of our users and our own.

Companions have the tendency to be exceedingly nice to each other. This is due to the data set used in fine-tuning the models and not related to the Drama Engine itself. Except it might be the case that all chat participants except the prompted companion in any situation are perceived as the user and thus the companions are acting as assistants to each other. New models that are trained for role-playing will be tested as they become available in the hope that more of the character of the companion gets preserved instead of its replies being overruled by the fine-tuning. Larger models show better role-playing capabilities \citep{wangRoleLLMBenchmarkingEliciting2023}.

Companions sometimes show a tendency of copying each other's conversation style. Again, this is most likely down to the size and fine-tuning of our most-used models. The longer the chat context, the less the companions are able to take their system prompt into account.

Trade-offs between model size and reply speed and quality. The above issues occur less if we use larger models but that comes with several downsides: inference cost, time, and resource usage goes up. We are very committed to using the least energy consuming model for every task and that means we will compromise quality.

We need models with better role-playing abilities. Because we do not have the infrastructure to create our own models, we will wait for those models to be developed by third parties. We are planning to evaluate the newly launched Hermes 3 (\citep{tekniumHERMESTECHNICALREPORT2024}) as a replacement for our current base model for this reason.

\section{Extending the Drama Engine}

The easiest way to extend the current functionality of the Drama Engine is to write new instruction deputies. Those just have a job defined that replaces most of the deputy prompt when sent to the model. The companion hosting the deputy then directly forwards the prompt’s result as if they had replied by themselves. Adding a new instruction deputy requires a line defining the deputy and another one defining an action that uses this deputy.

Further, users can write their own sub-classes of the \texttt{Deputy} class. In that case, all they have to do is create a new class, a companion config, and a \texttt{runAction} function. This function can edit the context to trigger automatic prompting of the language model. It can also just set a job or other context fields to specify the behaviour of the hosting companion instead. Automatically determining if an action should be triggered - instead of returning a standard reply - is something we are considering tackling as one of the next features for the Drama Engine, since the explicit action selection we’re supporting now has the upside of making the abilities of the system clear but the downside that it feels less natural than talking to a system with an “action model” in the background (e.g. Apple's Siri).

Additionally, reply functions allow for a wide range of customisation of companion behaviour all the way to full agentic workflows where one agent writes code, another one executes it and a third one criticises it. Or, the context object could be used to retrieve information from a knowledge graph, trans-code it to plain English and inject it into the companion prompt to create situational awareness. The sky's the limit.

\section{Future Outlook}

We are currently working on giving companions memory. Since we are already logging all conversations and have the ability to use an LLM for compressing them we can implement short-term as well as long-term memory.

Tool-calling is another interesting potential addition to the Drama Engine. At the moment we do not support it because of the lack of a unified syntax. It would be straight-forward to implement one of the existing solutions, e.g. following \citet{tekniumHERMESTECHNICALREPORT2024}, and generalise it if and when a standard emerges.

\section{Ethics Statement}
There are several potential ethical considerations that could arise from the development and use of the Drama Engine framework.

\begin{itemize}
    \item Privacy and Safety: The developer working with the Drama Engine is responsible for maintaining the safety and privacy of user communication with agents.
    \item Bias: Biases are present in model training data, resulting in biased model output \citep{navigliBiasesLargeLanguage2023}. While we encourage inclusivity and fairness, we acknowledge that artistic production requires the modelling of biases in order to give companions personality. This means that developers and designers have to be aware of the biases present and actively work with them in order to elicit the intended aesthetic manifestation that they seek.
    \item  Transparency: When companions reply to each other the conversation can take unexpected turns. In those cases, transparency is the best tool to further tune the behaviour of companions with the goal of minimising the problem. It has to be always clear to the end user that language models produce not fully predictable output.
    \item Reliability: While the Drama Engine is structured so that it provides a reliable interface to large language models, those models themselves are unreliable and so are hosting solutions (e.g. caching can lead to erroneous replies). It has to be always clear to the user that they are working with technology that carries a certain level of unpredictability with it. The intended use of the Drama Engine is entertainment and support – it should not be relied on in decision making situations that go beyond aesthetics.
\end{itemize} 

\section{Acknowledgements}
We want to thank Rockstart and Innofounder for their belief in us, the IT University of Copenhagen for supporting us on our first steps, and all of our users for coming on this ride with us.


\bibliographystyle{plainnat}
\bibliography{drama_engine.bib}

\end{document}